\documentclass[conference]{IEEEtran}
\IEEEoverridecommandlockouts
\usepackage[utf8]{inputenc}
\usepackage{bm}
\usepackage{algorithm2e}
\usepackage{multirow}
\usepackage{enumitem}
\usepackage{booktabs}
\usepackage{amsmath,amssymb,amsfonts}
\usepackage{algorithmic}
\usepackage{graphicx}
\usepackage{textcomp}
\usepackage{xcolor}
\usepackage{subcaption}
\RestyleAlgo{ruled}

\usepackage{cite}
\def\BibTeX{{\rm B\kern-.05em{\sc i\kern-.025em b}\kern-.08em
    T\kern-.1667em\lower.7ex\hbox{E}\kern-.125emX}}
\usepackage{url}

\title{Decentralized Personalized Online Federated Learning}
\author{
\IEEEauthorblockN{
    Renzhi Wu\textsuperscript{1*} \thanks{*Work done at Adobe Research.}
    \hspace{3ex}Saayan Mitra\textsuperscript{2}
    \hspace{3ex}Xiang Chen\textsuperscript{2}
    \hspace{3ex}Anup Rao\textsuperscript{2}
}
\IEEEauthorblockA{
    \textsuperscript{1}Georgia Institute of Technology, 
    \textsuperscript{2}Adobe Research \\
renzhiwu@gatech.edu, \{smitra, xiangche, anuprao\}@adobe.com
}
}

\begin{document}

\IEEEoverridecommandlockouts
\IEEEpubid{\makebox[\columnwidth]{978-1-5386-5541-2/18/\$31.00~\copyright2023 IEEE \hfill}
\hspace{\columnsep}\makebox[\columnwidth]{ }}

\maketitle
\IEEEpubidadjcol

\begin{abstract}
Vanilla federated learning does not support learning in an online environment, learning a personalized model on each client, and learning in a decentralized setting. There are existing methods extending federated learning in each of the three aspects. However, some important applications on enterprise edge servers (e.g. online item recommendation at global scale) involve the three aspects at the same time. Therefore, we propose a new learning setting \textit{Decentralized Personalized Online Federated Learning} that considers all the three aspects at the same time. 

In this new setting for learning, the first technical challenge is how to aggregate the shared model parameters from neighboring clients to obtain a personalized local model with good performance on each client. We propose to directly learn an aggregation by optimizing the performance of the local model with respect to the aggregation weights. This not only improves personalization of each local model but also helps the local model adapting to potential data shift by intelligently incorporating the right amount of information from its neighbors.  
The second challenge is how to select the neighbors for each client. We propose a peer selection method based on the learned aggregation weights enabling each client to select the most helpful neighbors and reduce communication cost at the same time. We verify the effectiveness and robustness of our proposed method on three real-world item recommendation datasets and one air quality prediction dataset.
\end{abstract}


\section{Introduction}
Machine learning, especially deep learning, requires large amounts of training data that is typically collected from user devices.
Traditionally, all user data is sent to and stored in a central server to perform model training. This practice has drawbacks in many aspects including privacy, security and latency. 
In particular, user data privacy has become a major concern in many machine learning applications~\cite{liu2021machine}.
Considering these drawbacks of centralized data collection/model training, Federated Learning~\cite{mcmahan2017communication, li2020review} has emerged to be a popular machine learning paradigm in which user data is kept on each local client (a regional edge server or even the user device) and a global model is trained in a distributed fashion under the coordination of a central server.

We consider federated learning on enterprise edge servers where each client is a regional edge server. 
In the rest of the paper, we will refer to "client" or "edge server" as "edge" following the convention in enterprise edge computing~\cite{shi2016edge}. 
In this scenario, user data from one region cannot (or will not be allowed to) be shared with edge/central servers in another region due to data governance laws and regulations (e.g. General Data Protection Regulation (GDPR)~\cite{voigt2017eu} in European Union, Personal Information Protection Law (PIPL)~\cite{determann2021china} in China, and California Consumer Privacy Act (CCPA)~\cite{goldman2020introduction}), which makes federated learning an attractive learning paradigm.
However, for some applications on enterprise edges,
vanilla federated learning suffers from three limitations.

Firstly, most federated learning methods are developed for offline training data~\cite{zhang2021survey, li2020review, li2020federated}. In this case, the whole training process happens only once or once in a while after the offline data is updated. 
However, in most real-world applications, the data arrives at the edges in an online fashion~\cite{hoi2021online}. In applications where the data distribution does not shift over time (or shift very slowly over time), one workaround is to accumulate some online data as offline data, then train the model with federated learning once in a while. 
This workaround is not feasible in applications where the data distribution can shift rapidly, in which case one has to continuously train the model in an online fashion to adapt the model to the dynamic data trends. One such application is recommendations in the fashion domain where new items/products are introduced very frequently and certain items can suddenly gain or lose in popularity due to random events (e.g. endorsement by a celebrity or scandalous news of a related company). 
In these applications, both learning and inference should also be performed online whenever new data points arrive. 
Specifically, the workflow is that, first 
some data points arrive (e.g. online browsing activities of an user), the model performs inference (e.g. select items to recommend), and feedback is received (e.g. whether the user has clicked on the recommended items.), then the model trains on the feedback. 
This process is continuously repeated online.
To the best of our knowledge, the only work in federated learning that considers the online setting is FedOMD~\cite{mitra2021online}. There are also several other methods titled "online federated learning"~\cite{chen2020asynchronous, damaskinos2020fleet, li2019online} but "online" is meant in a different way, e.g. clients join and drop online~\cite{li2019online}.

The second limitation is that vanilla federated learning aims at learning a single global model. However, in the case where the edges do not have the same data distribution, using a single global model is sub-optimal and it is better to have a personalized model on each edge that is adapted specifically for the data distribution on that edge. For example, having tailored models for the edge in France and the edge in the US is beneficial because their data distribution is different due to regional differences in culture.
To counter this limitation, different approaches have been attempted including federated multi-task learning~\cite{smith2017federated, corinzia2019variational, marfoq2021federated}, personalized layers~\cite{arivazhagan2019federated} and meta-learning based approaches~\cite{fallah2020personalized}. 

The third limitation is that the existence of the central parameter server (that coordinates the whole learning process) introduces a single point of failure
and regulatory risks. If the central server is down or the connection to the central server is congested, all edges will be impacted. 
Furthermore, centralized parameter servers are less flexible regarding different and changing regulations. For example, some countries (e.g. China) suddenly disallow direct model parameter sharing (or limit network connection) to the country where the centralized parameter server is located (e.g. the US), then the centralized parameter server has to move to a different place to conform to the regulation. The same situation may happen multiple times, so in an extreme scenario, there might be no suitable country to host the centralized parameter server. However, this situation can be easily handled in a decentralized setting where the local edges (servers) in China can simply disable the links to the edges (servers) in the US to conform to regulations.

Therefore, decentralized federated learning has been considered~\cite{lalitha2018fully, roy2019braintorrent, hu2019decentralized} where each edge can directly communicate with other edges. 
One challenge of decentralized federated learning is peer selection, which is important but under-explored. The motivation of peer selection is that if every edge has to exchange model parameters with every other edge, the amount of communication is $O(n^2)$ where $n$ is the number of edges. 
To reduce communication cost, existing work typically use simple heuristics, e.g., by hard-coding each edge/client to communicate only with certain peers~\cite{lalitha2018fully} or randomly sampling some peers to connect~\cite{hu2019decentralized}. 
Intuitively, it would be the best if each edge/client avoids the peers that can be "adversarial" and selects the peers that are most helpful to it, especially in the case where the edges do not have the same distribution and we want to learn a personalized model for each edge. However, peer selection strategies smarter than hard-coded rules and random sampling remain unexplored.

As discussed, there has been existing work focusing on tackling each one of the three limitations.
However, some important applications (that are at a global scale) involve all three challenges at the same time. For example, fashion/news item recommendations for users;  Propensity analysis of customers performing certain actions; Personalized discount offerings; Real-time bidding of ads; Trend detection in e-commerce/social media. Each of these applications is ubiquitous in most tech companies that operate on a global scale and involves billions of users. 

We propose a new setting: \textit{decentralized personalized online federated learning} to consider the three limitations at the same time. Specifically, "decentralized", "personalized", and "online" target the three aforementioned limitations respectively. To the best of our knowledge, this setting has not yet been considered in literature. A straightforward solution for this setting is to directly combine existing methods developed specifically for each limitation, which however is infeasible as the techniques are mostly not compatible.
Therefore, we initialize the study on decentralized personalized online federated learning by proposing a simple yet effective method. 

To do online learning, we consider only the type of models that are trained with online stochastic gradient descent (SGD); 
To perform decentralized federated learning to incorporate information from other edges, each edge periodically fetches model parameters from other edges and aggregates them with its local model. The core idea of our method is about how to perform aggregation. In existing federated learning methods, the aggregation weights are hard-coded\cite{mcmahan2017communication, DBLP:conf/iclr/LiHYWZ20}. We propose to learn a set of personalized aggregation weights for each edge so that the aggregated model is personalized for the edge by allowing each edge to intelligently select the amount of information to incorporate from its neighbors. The learned aggregation weights are also dynamically adapted over time which helps each edge adapting to data shift over time. 
We then further exploit the learned aggregation weights to perform intelligent peer selection where the intuition is that there is no need for one edge to communicate with another edge that would have a very small weight during aggregation. 

To summarize, we make the following contributions:
\begin{itemize}[leftmargin=*]
    \item We consider a new learning setting, decentralized personalized online federated learning, which targets at applications that require reliability, personalization, online learning, and privacy at the same time. 
    \item We propose to dynamically learn a set of personalized aggregation weights for each local model directly through gradient descent. The learned aggregation weights allow each edge to incorporate the right amount of information from its neighboring edges and also improves personalization of each edge.
    \item Exploiting the learned aggregation weights, we design a greedy peer selection method for each edge to select the most helpful set of peers as its neighbors to communicate with, reducing communication cost.
    \item We perform extensive experiments with item recommendation and regression tasks on real-world data to verify the effectiveness and robustness of our method. 
\end{itemize}

\section{Related Work}
Since the existing work focuses on each one of the three mentioned limitations of federated learning, the most relevant work to us includes online federated learning, personalized federated learning, and decentralized federated learning. The major difference of our method from these existing methods is that we consider the three limitations at the same time. 
We discuss the differences/similarities of our method from/to the existing methods in each aspect (online learning, personalized learning, and decentralized learning) in the following. 

\noindent\textbf{Online federated learning.} 
To the best of our knowledge, the only work in online federated learning is FedOMD~\cite{mitra2021online} that incorporates ideas from online learning into federated learning. FedOMD considers the case when the loss is convex or strongly convex and proposes a mirror descent based method that enjoys good regret bounds. However, most real-world applications involve non-convexity, so the application of FedOMD is limited.
There are several other works titled "online federated learning"~\cite{chen2020asynchronous, damaskinos2020fleet, li2019online} but "online" is meant in a different way, e.g. clients join and drop online~\cite{li2019online} or the model is first trained on online data for some time (without doing any inference) and after that the model performs inference on the successive online data (without doing any training)~\cite{chen2020asynchronous}. 
Similar to~\cite{chen2020asynchronous}, we also use online stochastic gradient descent (SGD) 
to train our models. However, the difference is that we follow the standard online learning setting~\cite{shalev2012online} (as in FedOMD~\cite{mitra2021online}) where we perform inference on each batch of newly arrived data points and then receive the ground-truth feedback of the data to train the model. 

\noindent\textbf{Personalized federated learning.} 
There are several types of methods to learn a personalized model on each edge (client) in federated learning~\cite{kulkarni2020survey} including federated multi-task learning~\cite{smith2017federated, corinzia2019variational, marfoq2021federated}, personalized layers~\cite{arivazhagan2019federated}, meta-learning based methods~\cite{fallah2020personalized}, and mixture of local and global model~\cite{deng2020adaptive}. The most relevant method to us is the mixture of local and global model where during aggregation the central server seeks a balance between every local model and the global model. At high level, it first obtains a global model $\bar{M}$ by aggregating all local models $M^1, \dots, M^n$, after that, it produces a personalized model $(1-\lambda)\bar{M}+\lambda M^i$ for the $i^{th}$ edge where $\lambda$ is a hyper-parameter controlling the balance between the global model $\bar{M}$ and the local model $M^i$. 
In our method, we similarly seek a balance between an edge and its neighboring edges during model aggregation. The differences are that firstly, in our method aggregation is done on each edge instead of the central server as we consider a decentralized setting; secondly, on each edge during aggregation we directly learn a balance (i.e. the aggregation weights for the local and neighboring models) with the upcoming local data instead of using a hyper-parameter $\lambda$ that is difficult to tune in practice. 

\noindent\textbf{Decentralized federated learning.}
Existing work has focused on the communication infrastructure for decentralized federated learning for example braintorrent~\cite{roy2019braintorrent} and 
blockchain-based methods~\cite{li2020blockchain} or  better training procedures of the models~\cite{lalitha2018fully}.
A key component in decentralized federated learning methods is a strategy to select peers as neighbors for each edge to communicate with (to reduce the amount of communication). 
However, peer selection is an under-explored problem as existing strategies are either hard-coding (e.g. selecting one-hop neighbors only)~\cite{lalitha2018fully,jiang2021asynchronous, hegedHus2019gossip} or random sampling~\cite{roy2019braintorrent, hu2019decentralized, pappas2021ipls}.
Peer selection is especially important for learning personalized models, as each edge would like to avoid potential "adversarial" peers and select the peers that are the most helpful.
In our method, we propose a peer selection strategy (that is smarter than hard-coding and random sampling) based on the learned aggregation weights.  

\section{Preliminaries}
We briefly introduce the formulations of Federated Learning and Online Learning that serve as the building blocks for decentralized personalized online federated learning. 

\noindent\textbf{Federated Learning.} There is one central server and $n$ edges $e^1, \dots, e^n$
where each edge has a local model $M_t^i$ at timestamp $t$. Note we use the notation $M$ to refer to a model and also the parameters of the model.
Federated learning repeatedly executes the following two steps until convergence:
\begin{itemize}[leftmargin=*]
    \item \textit{(1) Local model learning}. Each edge $e^i$ downloads the global model from the central server and trains the global model on its local data which becomes its local model $M^i$.
    \item \textit{(2) Centralized model aggregation.} The central server receives updated local models from all edges and aggregate all received local models to build a global model. 
\end{itemize}
Most federated learning methods use weighted average of local model parameters during aggregation in step (2). For example, in the popular FedAvg algorithm~\cite{mcmahan2017communication}, the global model $\bar{M}_t$ is obtained by $\bar{M}_t=\frac{\sum_{i=1}^n \alpha^i M^i_t}{\sum_{i=1}^n \alpha^i}$ where aggregation weight $\alpha^i$ is set to be the amount of data on the edge $e^i$.

\noindent\textbf{Online Learning.}
Online learning refers to the following learning setting~\cite{shalev2012online}: At every time stamp $t$, a new batch of data $X_t$ arrives, the model make predictions $\hat{y}_t=M_t(X_t)$, and then the ground-truth $y_t$ is revealed; The model learns from $(X_t, y_t)$ and is updated as $M_{t+1}$; This process is repeated whenever a new batch of data arrives.
It is also possible that the ground-truth is revealed with a time delay $\delta t$. In this case, similarly, at time stamp $t$, the model $M_t$ learns from $(X_{t-\delta t}, y_{t-\delta t})$ and becomes $M_{t+1}$. 

Most existing literature on online learning focuses on the convex setting~\cite{hoi2021online, shalev2012online, mitra2021online}. For neural networks, a popular method is to use online stochastic gradient descent (SGD)~\cite{chen2020asynchronous}. Specifically, $M_{t+1}$ is obtained from $M_{t}$ by taking one or few gradient descent steps on data $(X_t, y_t)$. 

We note that online learning assumes ground-truth is revealed soon after prediction is made. This is the case in many important applications, for example, stock price prediction, weather prediction, and online item recommendation.

\section{Proposed Learning Setting}
To resolve the three limitations of vanilla federated learning at the same time, we propose a new learning setting \textit{Decentralized Personalized Online Federated Learning}.

Each edge $e^i$ has a local model $M^i$ which is learned \textbf{online}. Specifically, when on edge $e^i$ at timestamp $t$, a new batch of data $X_t^i$ arrives; the current model $M_t^i$ makes prediction $\hat{y}_t^i$ for the data batch and then receives feedback $y_t^i$ which serves as the ground truth. The model $M^i_t$ learns on the new data $(X_t^i, y_t^i)$ and evolves to $M^i_{t+1}$.

Let $\mathcal{E}^i_t$ denote the set of indices of edges that are the neighbors of edge $e^i$ at timestamp $t$.
In the most simplistic case, each edge chooses all other edges to be its neighbor, then $\mathcal{E}^i_t= \{j|j\neq i; 1\leq j\leq n\}$. 
The edges are \textbf{decentralized} and each edge directly communicates with its neighbors in a peer-to-peer fashion.  Note neighbor means logical unidirectional neighbor, i.e. neighbors do not have to be physically close and it is possible that one edge includes another edge as neighbor but not vice versa. 

Each edge is allowed to fetch the model parameters from its neighbors to incorporate information of the neighboring edge without sharing user data as in \textbf{federated} learning. For each edge $e^i$, after every $T_{\text{agg}}$ timestamps, it fetches models from its neighbors and aggregates its local model with the fetched models. Formally, it fetches $\{M^j|j\in \mathcal{E}^i_t\}$ from all neighbors and aggregates $\{M^j|j\in \mathcal{E}^i_t\}$ with its local model $M_{t}^i$:
\begin{equation}
    M_{t, \text{agg}}^i = g_{\bm{\alpha}_t^i}( \{M^j|j\in \mathcal{E}^i_t\}, M_{t}^i)
\end{equation}
where $g_{\bm{\alpha}_t^i}()$ is an aggregation function parameterized by $\bm{\alpha}_t^i$. Note the models $M^j$ do not have a subscript $t$ as the models from other edges may not to be synchronized, i.e. for $j\neq i$, $M^j$ can be from timestamp $t'$ with $t'\neq t$. The set of neighbors $\mathcal{E}^i_t$ and the aggregation parameter $\bm{\alpha}^i_t$ can change over time.

The set of neighbors $\mathcal{E}^i_t$ and the aggregation parameter $\bm{\alpha}^i_t$ of the aggregation function are both specific or \textbf{personalized} to each edge $e^i$ meaning that each edge can integrate information from its preferred edges in customized ways. This allows each edge to obtain a personalized model. 

\begin{figure*}[!htb]
    \centering
    \includegraphics[width=17.5cm]{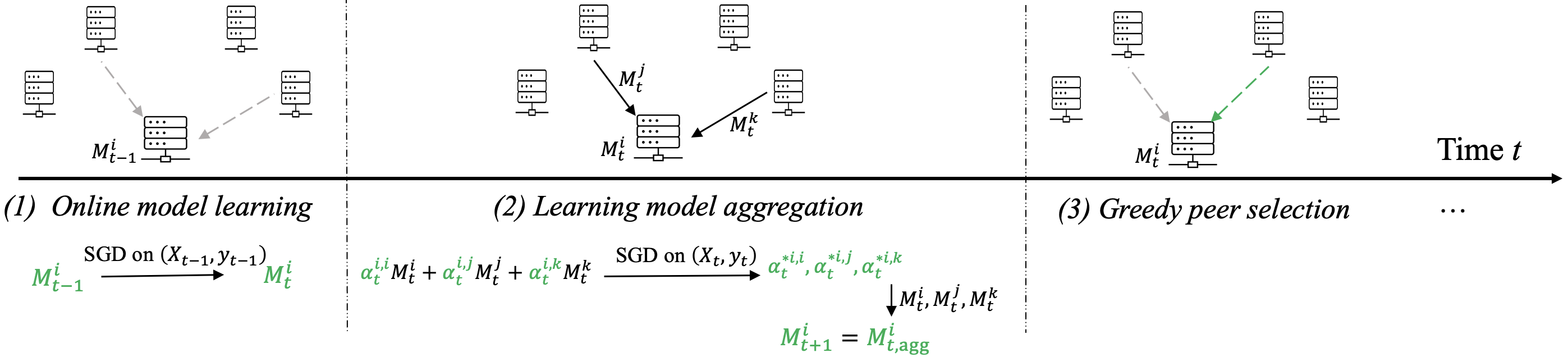}
    \caption{Overall workflow of our method on one edge ($e^i$). The three steps happen continuously and repeatedly.}
    \label{fig:workflow}
    \vspace{-7mm}
\end{figure*}

\noindent\textbf{Design Space.} The above specifications define our proposed learning setting. The design space for our setting includes: 
\begin{itemize}[leftmargin=*]
    \item  How to perform online learning? Specifically, how does model $M^i_t$ learn from data $(X_t^i, y_t^i)$ to get $M^i_{t+1}$? 
    \item How to select the set of neighbors $\mathcal{E}^i_t$ for each edge $e^i$ at each timestamp $t$? 
    \item How to perform aggregation for each edge? i.e. what is the form of $g_{\bm{\alpha}^i_t}()$ for each edge $e^i$ at each timestamp $t$?
\end{itemize}

\section{Proposed Method}
To derive our method, we first decide on how to perform online learning. As discussed, online stochastic gradient descent (SGD) is commonly used for neural networks~\cite{chen2020asynchronous}. Therefore, we perform online learning through online SGD. 

For model aggregation and peer selection, we introduce an unified solution. 
Our idea is to use weighted average aggregation where the aggregation weights for neighboring edges are directly learned specifically for each edge. 
Each edge can then select peers with higher weights to be its selected neighbors in the next round. The workflow of our method at high level is shown in Figure~\ref{fig:workflow}. First, each model performs online learning with online SGD independently. Second, when aggregation is invoked for an edge $e^i$, it fetches model parameters from its neighbors (e.g. it has two neighbors $e^j$ and $e^k$), and figures out the aggregation weights $\alpha_t^{*i,i}$, $\alpha_t^{*i,j}$, and $\alpha_t^{*i,k}$ 
for its local model and the models from neighbors.
The aggregated model $M_{t, \text{agg}}^i$ is then obtained with the weights. Third, each edge $e^i$ updates its set of neighbors based on the learned aggregation weights; Intuitively, neighbors with very small aggregation weights should be replaced, as including them makes little difference to the aggregated model. This process is continuously repeated.  We introduce the details next.%

\subsection{Learned aggregation}
We adopt the weighted average aggregation (which is also commonly used in exising federated learning methods~\cite{mcmahan2017communication, DBLP:conf/iclr/LiHYWZ20}), formally at edge $e^i$:
\begin{equation}
    M_{t, \text{agg}}^i =g_{\bm{\alpha}_t^i}( \{M^j|j\in \mathcal{E}^i_t\}, M_{t}^i)= \frac{\alpha_t^{i,i}M_t^i+\sum_{j \in \mathcal{E}^i_t}\alpha_t^{i,j}M^j}{\alpha_t^{i,i} + \sum_{j \in \mathcal{E}^i_t}\alpha_t^{i,j}}
\end{equation}
where $M^j$ represents the model parameters of the model received from the edge $e^j$; $\bm{\alpha}_t^i = \{\alpha_t^{i,i}\} \cup \{\alpha_t^{i,j}|j\in \mathcal{E}^i_t\}$ is the aggregation weight vector on edge $e^i$ in which each $\alpha_t^{i,j}$ is a scalar value representing the weight of model $M^j$ during aggregation on edge $e^i$.

In existing federated learning methods, aggregation weights are set with fixed heuristic rules e.g.  relating to the difference of model weights~\cite{huang2021personalized, zhang2020personalized} or set to be the amount of data on each edge~\cite{mcmahan2017communication, DBLP:conf/iclr/LiHYWZ20}. 
Inutively, these fixed rules could be suboptimal. 
As an example, we illustrate the suboptimality of the most commonly used method, i.e. setting weights to be the amount of data on each edge.
Consider the case there are three edges $e^i$, $e^j$ and $e^k$; Let's say $e^i$ and $e^j$  have similar data distribution, while $e^k$ has a large amount of data but a completely different data distribution from the two other edges. We focus on aggregation on edge $e^i$. 
With the strategy of setting weights to be the amount of data on each edge the weight $\alpha_t^{i,k}$ for the model from $e^k$ would be the greatest which is clearly sub-optimal. 
Instead, it might be better to have a larger weight $\alpha_t^{i,j}$ for the model from $e^j$, as the data distribution is similar.   
Since the data distribution as well as the amount of data on different edges can shift over time in the online learning environment, we also want the weights to be able to be dynamically adapted according to data shift. 

In contrast, we propose to directly learn the aggregation weights directly through gradient descent by exploiting the online nature of our setting. 
Without loss of generality, we focus on edge $e^i$.
Our idea is that in our online learning setting, when timestamp $t$ involves aggregation, we use the ground-truth feedback $y_t^i$ of data batch $X_t^i$ to learn the aggregation weights instead of using the data batch to train the local model.

Specifically, Let $M_{t, \text{agg}}^i(X_t^i)$ denote the prediction of aggregated model $M_{t, \text{agg}}^i$ on data batch $X_t^i$. Recall that $M_{t, \text{agg}}^i = \frac{\alpha_t^{i,i}M_t^i+\sum_{j \in \mathcal{E}^i_t}\alpha_t^{i,j}M^j}{\alpha_t^{i,i} + \sum_{j \in \mathcal{E}^i_t}\alpha_t^{i,j}}$. We freeze each model (each $M^j$ and $M_t^i$)  and only keep the aggregation weights $\bm{\alpha}_t^i = \{\alpha_t^{i,i}\} \cup \{\alpha_t^{i,j}|j\in \mathcal{E}^i_t\}$ as the active parameters. We then minimize the loss function $L(M_{t, \text{agg}}^i(X_t^i), y_t^i)$ with respective to aggregation weights $\bm{\alpha}_t^i$ with gradient descent. This provides us a set of learned aggregation weights $\bm{\alpha}_t^{*i}$. Formally:
\begin{equation}
\label{eq:learn_agg_weights}
    \bm{\alpha}_t^{*i} = \arg\text{min}_{\bm{\alpha}_t^{i}} L(M_{t, \text{agg}}^i(X_t^i), y_t^i)
\end{equation}
Once the aggregation weights are learned, we obtain an aggregated model $M_{t, \text{agg}}^i$ with the learned weights. The aggregated model is then used as the local model $M_{t+1}^i$. 

This process is also depicted in Figure~\ref{fig:workflow} where green color highlights items that are active. During online learning in step (1), the model is active so gets updated when performing SGD. During model aggregation in step (2), the only aggregation weights are active and they gets updated with SGD.

\noindent\textbf{Handling Overfitting.}
When the batch size is small,
Equation~\ref{eq:learn_agg_weights} can cause overfitting. 
We can pause learning (but inference continues) for several data batches to accumulate enough data for learning the aggregation weights. 
Nevertheless, minimizing the loss with respect to the current batch of data has the risk of overfitting the current batch of data. To prevent over-fitting, we learn the aggregation weights also with online gradient descent. Specifically, when we do aggregation at time stamp $t$, we initialize the weights as the weights learned from the previous aggregation step, and then update the weights with a only few gradient descent steps. This is repeated in every aggregation step.

\noindent\textbf{Reducing Inference Latency.} Since learned model aggregation happens online, one concern is how it affects inference latency. To make sure inference latency is not impacted by learned model aggregation, one can use a duplicate model in parallel to do learned aggregation. 
Specifically, when aggregation starts, a duplicate model $M_\text{dup}$ is created. Learned aggregation happens on $M_\text{dup}$, while at the same time, inference is done with the original model $M$ in parallel. When $M_\text{dup}$ finishes training, the original model $M$ is replaced by $M_\text{dup}$ which can be done efficiently.

\subsection{Greedy Peer Selection}
\label{sec:peer_sampling}
When the number of edges $n$ is large, if each edge includes all other edges as its neighbors, the amount of communication during aggregation is $O(n^2)$ which can be huge. 
To reduce communication cost, the common practice in decentralized learning is that each edge select $K$ other edges as its neighbors to communicate with so that the amount of communication is reduced to $O(nK)$. 

Intuitively, each edge should select the $K$ edges that are the most helpful to it as its neighbors, which is however difficult to achieve. 
There are two strategies to select edges/peers in existing work on decentralized federated learning. The first strategy is to simply hard-code the neighbors for each edge (e.g. selecting the physical one/few-hop neighbors for each edge)~\cite{lalitha2018fully,jiang2021asynchronous, hegedHus2019gossip}.
The second strategy is to randomly sample edges as neighbors~\cite{roy2019braintorrent, hu2019decentralized, pappas2021ipls, jelasity2007gossip, jelasity2004peer}. 
To the best of our knowledge, no existing work in decentralized federated learning has attempted a peer selection approach that is more intelligent than hard-coding or random sampling. 

We propose a peer selection approach that is more intelligent than random sampling. The core intuition of our approach is that, for one edge $e^i$, if during aggregation the learned aggregation weight of its neighbor $e^j$ is very small or even equals to zero (i.e. $\alpha_t^{i,j}=0$), it might be helpful to replace the neighbor $e^j$ with another edge $e^k$. More generally, we will be able to drop or select peers for each edge based on the learned aggregation weights. We note that this is meaningful only when the aggregation weights are changing like in our work. In all existing work, the aggregation weights are not dynamically learned and so that it was not possible to intelligently select peers based on aggregation weights.

Without loss of generality, we consider edge $e^i$. Ideally, if we have access to the learned aggregation weights for every other edge, we can just keep the top $K$ peers with the highest weights. However, the weights are unknown unless $e^i$ communicates and receives the models from every other edge, which is exactly what we want to avoid in the first place. Therefore, we propose a greedy approach that greedily replaces the neighbor of the smallest weights with the most "promising" unconnected edges based on the observed aggregation weights. The method has the following steps:
\begin{enumerate}[leftmargin=*]
    \item Initially, edge $e^i$ randomly select $K$ peers as its neighbors $\mathcal{E}^i_t$. (Similar for every other edge) %
    \item After, $m$ aggregation steps, we have learned the aggregation weight vector $\bm{\alpha}_t^i$ for the neighbors of $e^i$. We first normalize the weight vector (e.g. $\bm{\alpha}_t^i$) to make it sum to 1 for each edge and then  greedily obtain the "weight" for each two-hop neighbor $e^k$ of $e^i$ by:
\begin{equation}
    \alpha^{i,k}_t = \sum_{j \in \mathcal{E}^i_t, \text{ if } k\in \mathcal{E}^j_t} \alpha^{i,j}_t\alpha^{j,k}_t
\end{equation}
The two-hop neighbors with the highest weights are intuitively the most "promising" unconnected peers for $e^i$.
\item We replace the $K'$ peers in $\mathcal{E}^i_t$ that have the smallest aggregation weights with the $K'$ most "promising" unconnected peers (two-hop neighbors) for $e^i$.
We initialize the aggregation weights for the newly added neighbors to be zero.
\item We continuously repeat step (2) and step (3), to enable continuously exploration. 
\end{enumerate}

With the above procedures, we are able to dynamically select peers as neighbors for each edge. To enable each edge to add and drop neighbors easily (especially in case of adding brand new edges to the network or removing edges from the network), we can simply
maintain a server for broadcasting the overall topology (or connectivity) to all edges, which is allowed as no user data is involved, or one can adopt existing methods used in network routing~\cite{medhi2017network}.

\noindent\textbf{Optimizing Communication for Id Embeddings.} For typical models like MLP, the whole model is fetched from a neighbor for each edge. For models (e.g. deepFM~\cite{guo2017deepfm}) with embeddings for "ids" like "userid" and "itemid", different edges may have little overlap on the these "ids". Therefore, to minimize communication cost, when fetching the model from a neighbor, each edge only fetch the embeddings for common "ids". In addition, in practice, most id embeddings stay unchanged between two aggregation steps, for example, the "userid" embeddings for the users that are inactive during two aggregation steps do not change. Therefore, each edge only fetch the embeddings for the common "ids" that are updated since the last aggregation step. This optimization reduces communication cost a lot for models like deepFM~\cite{guo2017deepfm}. 

\subsection{Discussion}
Similar to most existing federated learning settings, our proposed setting also shares model parameters, which has drawbacks in privacy. Specifically, the model parameters are learned from user data, i.e. model parameters encode patterns in user data, so it is possible to do reverse-engineering to recover some information about the user data from the model parameters~\cite{phong2017privacy, hitaj2017deep}. We note that this limitation is not unique to our setting and one could adapt existing solutions developed for federated learning (e.g. encryption-based approaches~\cite{aono2017privacy, bonawitz2017practical} and differential privacy-based approaches~\cite{wei2020federated, triastcyn2019federated}) to our setting to counter the issue. In this work, we consider this issue of indirect leakage as an orthogonal problem and leave it to future work.

\section{Experiments}
We evaluate our method in the following dimensions:
\begin{itemize}[leftmargin=*]
    \item How does the performance of each of our proposed component compare to that of the alternative solutions?
    \item How robust is our method to different types of variations (e.g. unstable network connection)?
    \item How do the methods behave under different hyperparameters?
\end{itemize}

\begin{figure*}[!htb]
    \centering
    \includegraphics[width=17.5cm]{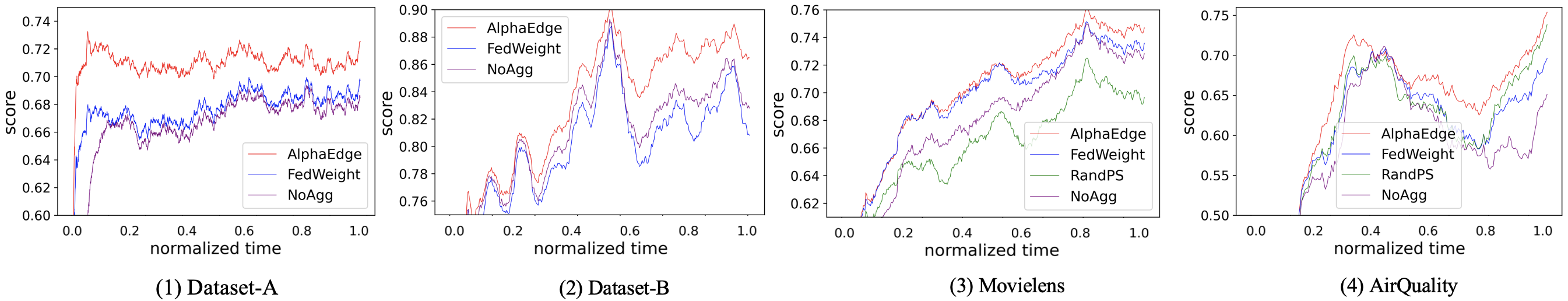}
    \vspace{-2mm}
    \caption{Scores over time on one typical edge. RandPS is identical with AlphaEdge on Dataset-A and Dataset-B.}
    \vspace{-7mm}
    \label{fig:scores_vs_t}
\end{figure*}

\subsection{Setup}
\noindent\textbf{Datasets:} 
We note that standard datasets (e.g. CIFAR-10 and MNIST) are typically homogeneous (and do not have timestamp attached), so they do not require personalized (and online) models. 
Therefore, we consider two private datasets from industry applications and one open dataset (which are not homogenous), and then adapt a standard datasets (movielens).

\begin{itemize}[leftmargin=*]
    \item \textit{Dataset-A.} This is a private dataset of user-item interactions (view vs purchase) from an online store. The items have attributes including category, price, inventory, etc. 
    Each user-item interaction record is associated with a "edge" attribute representing the edge server the record is from. Each edge server is at a different region. There are 4 edges and about 40 million user-item interaction records in total. Information of each item is available in a catalog table with attributes like price, size and inventory which we use as additional features. 
    \item \textit{Dataset-B}. This is a private dataset of user-item interactions (view vs purchase) from another online store. This dataset has the same format as Dataset-A, i.e. one table with user-item interaction records and one table with information about the items. There are 4 edges and about 20 million user-item interaction records in total.
    \item \textit{Movielens}~\cite{harper2015movielens}. This is a movie ratings dataset. 
    Following prior work~\cite{zhou2018deep}, we adapt it to be a Clickthrough rate (CTR) prediction task by binarilizing the ratings. 
    Since there is no edge in this dataset, we group the dataset into ten edges/regions according to the first digit of the zipcode for each record.
    We use Movielens-1m as it is the only movielens dataset that contains zip-code information. 
    There are 10 edges and about 1 million user-item interaction records in total. Since there is no distributional different between edges, in which case personalized model is not needed, so we inject label noise (by randomly flipping 10\% of the labels) to one edge to simulate distribution shift. 
    \item \textit{AirQuality}~\cite{airquality, chen2020asynchronous}. This is a dataset about predicting air quality in Bejing. The features are air quality and weather conditions (e.g. wind and precipitation) in the past, and the task is to predict the air quality (PM2.5) in the next hour. There are data from 35 weather stations and each weather station has its hourly measurement of the weather and air quality. We treat each weather station as one edge, so there are 35 edges. There are 0.4 million records in total.
\end{itemize}
All datasets have a timestamp column, so we simulate data record arriving batch by batch based on the timestamp column. 

\noindent\textbf{Base models:}
Our method is a wrapper method supporting aggregation of different base models on different edges. To perform experiments, we need to use a specific base model for the tasks. 
For the item recommendation datasets (Dataset-A, Dataset-B, and Movielens), we use deepFM~\cite{guo2017deepfm} as it is a popular method and achieves top performance on several benchmarks~\cite{deepfm_site}. For the regression task (AirQuality) we use a Multi-layer Perception (MLP) regressor, as it is commonly used in air quality prediction~\cite{ durao2016forecasting, peng2015air, garcia2018air}.

\noindent\textbf{Evaluation:}
We evaluate the methods based on inference performance sequentially on each batch of data.  
Specifically, when a new batch of data comes on one edge, the model (deepFM or MLP) first performs prediction on which the model's performance is evaluated (after that, the model receives and learns the ground-truth labels). We use AUC score (the area under the ROC curve) for the item recommendation datasets and 1- SMAPE score (Symmetric mean absolute percentage error~\cite{SMAPE}) for the regression dataset. For both metrics, the score value is in region [0, 1] and a higher score means better performance. %

\noindent\textbf{Methods:}
Since our work is the first work in decentralized personalized online federated learning, we are not able to find any baseline that is in the same setting. 
Therefore, we try to adapt existing methods to our setting. 

To the best of our knowledge, the only work in federated learning that considers the online setting is FedOMD~\cite{mitra2021online}. However, FedOMD considers convex scenario which is not applicable to our considered real-world dataset and code is not available. 
There are also several other methods titled "online federated learning"~\cite{chen2020asynchronous, li2019online} but "online" is meant in a different way, e.g. clients join and drop online~\cite{li2019online}.
Therefore, we consider adapting from standard federated learning~\cite{mcmahan2017communication, kairouz2021advances} and decentralized online learning~\cite{wan2022projection, jiang2021asynchronous} methods.
\begin{itemize}[leftmargin=*]
    \item \textit{AlphaEdge}. This is our proposed method, which includes our proposed two components: learned aggregation weights and greedy peer selection.
    \item \textit{FedWeight}. 
    This is a method adapted from standard federated learning~\cite{mcmahan2017communication, kairouz2021advances} by considering an online and decentralized scenarios. Its aggregation weights are related to the amounts of data on each edge.
    \item \textit{UniWeight}. 
    This is adapted from a decentralized online learning method~\cite{jiang2021asynchronous}.
    Its aggregation weights are uniform weights (an assumption often used in these literature). 
    \item \textit{RandPS}. This is to replace our greedy peer selection component with random peer selection at every aggregation step. This is commonly used in existing work~\cite{roy2019braintorrent, hu2019decentralized, pappas2021ipls}. 
    \item \textit{NoPS}. This is to remove peer sampling from AlphaEdge, i.e. each edge device communicate with every other edge device involving quadratic number of communications.
    \item \textit{NoAgg}. This method trains the model on each edge independently and there is no communication between the edges, i.e. it is to use the base model deepFM~\cite{guo2017deepfm} on each edge. 
    This is also equivalent to dropping both learned aggregation weights and greedy peer selection from AlphaEdge.
\end{itemize}

\noindent\textbf{Implementation and Hyperparameters.} 
We use the Adam optimizer~\cite{kingma2014adam} with a learning rate of $0.001$ for learning the local models as well as the aggregation weights, as the learning rate works well for deepFM. 
When simulating the arriving data batch stream, we use a batch size of $B=500$ for the two larger datasets Dataset-A and Dataset-B and a batch size of $B=50$ for the two smaller datasets Movielens and AirQuality. 
For each data batch, the local model performs one online SGD step with a batch size of $B$.
We use the binary cross entropy loss for the item recommendation datasets and mean squared error loss for the regression dataset. 
We perform aggregation after every $E=20$ local epochs (SGD steps or data batches).
When learning the aggregation weights with online SGD, we take $10$ SGD steps during aggregation. 
For peer selection, we set $K=5, K'=1$ and $m=1$ for every dataset, i.e. each edge has $K=5$ neighbors and in our method at every $m=1$ aggregation step we explore $K'=1$ promising peer by replacing the neighbor with the least aggregation weights. We report results averaged from five runs. 

\noindent\textbf{Hardware:} All experiments are performed on a machine with 8 v100 GPUs, 64 vCPUs, and 400 GB memory.

\begin{figure*}[!htb]
    \centering
    \includegraphics[width=17.5cm]{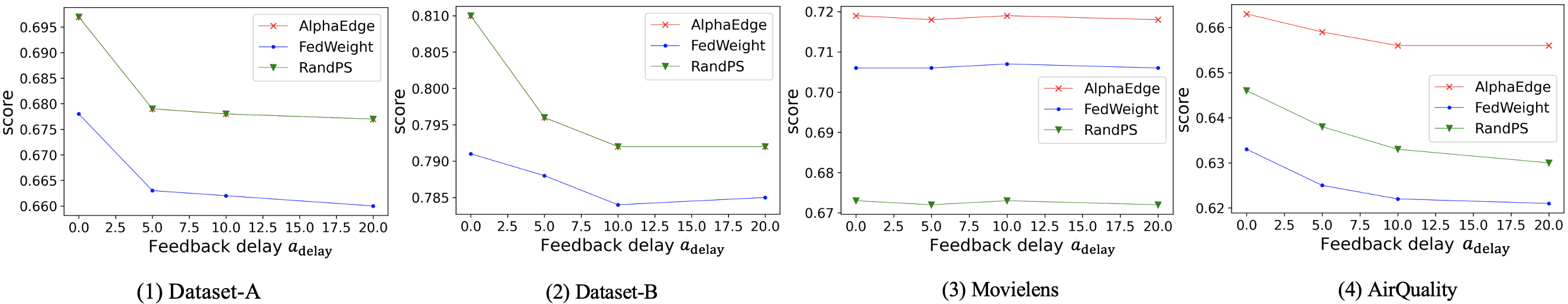}
    \vspace{-2mm}
    \caption{Averaged score vs feedback delay $\alpha_{\text{delay}}$. RandPS is identical with AlphaEdge on Dataset-A and Dataset-B.}
    \vspace{-7mm}
    \label{fig:delay}
\end{figure*}

\subsection{Overall Performance}

\begin{table}[!htb]
\vspace{-3mm}
\centering
\setlength{\tabcolsep}{1pt}
\caption{Scores of all methods on all datasets.}%
\begin{tabular}{|l|l|ll|ll|l|}
\hline
\multirow{2}{*}{} & \multirow{2}{*}{NoAgg} & \multicolumn{2}{l|}{\begin{tabular}[c]{@{}l@{}}Aggregation weights\end{tabular}} & \multicolumn{2}{l|}{\begin{tabular}[c]{@{}l@{}}Peer selection\end{tabular}} & \multirow{2}{*}{AlphaEdge} \\ \cline{3-6}
                  &                         & \multicolumn{1}{l|}{FedWeight}                         & UniWeight                        & \multicolumn{1}{l|}{RandPS}                          & NoPS                            &                     \\ \hline
Dataset-A          & 0.671                   & \multicolumn{1}{l|}{0.678}                          & 0.637                          & \multicolumn{1}{l|}{\textbf{0.697}}                  & \textbf{0.697}                  & \textbf{0.697}      \\ \hline
Dataset-B          & 0.796                   & \multicolumn{1}{l|}{0.791}                          & 0.758                          & \multicolumn{1}{l|}{\textbf{0.810}}                  & \textbf{0.810}                  & \textbf{0.810}      \\ \hline
Movielens         & 0.699                   & \multicolumn{1}{l|}{0.706}                          & 0.696                          & \multicolumn{1}{l|}{0.673}                           & \textbf{0.719}                           & \textbf{0.719}      \\ \hline
AirQuality        & 0.631                   & \multicolumn{1}{l|}{0.633}                          & 0.638                          & \multicolumn{1}{l|}{0.646}                           & 0.657                           & \textbf{0.663}      \\ \hline
Avg.              & 0.699                   & \multicolumn{1}{l|}{0.702}                          & 0.682                          & \multicolumn{1}{l|}{0.707}                           & 0.720                           & \textbf{0.722}      \\ \hline
\end{tabular}
\label{tbl:performance}
\vspace{-3mm}
\end{table}

The scores of all methods on all datasets are shown in Table~\ref{tbl:performance}.

Note it does not make much sense to do peer selection when the total number of edges is small (e.g. four edges for Dataset-A and Dataset-B and each edge only has three neighbors). We set $K=5$ so that peer selection only happens on Movielens and BeijingAir. 
On Dataset-A and Dataset-B, since we are allowing 5 neighbors, RandPS and NoPS are the same as AlphaEdge.

Our method AlphaEdge works the best. AlphaEdge has better performance than NoAgg because aggregating the models from other edges incorporates more information which helps with the local model performance. 

Replacing the learned aggregation weights with existing methods of choosing the weights decreases performance to be about the same as NoAgg. This is because blindly aggregating models can some times help (e.g. on Dataset-A) if the data distribution is similar  and can some times hurt (e.g. on Dataset-B) is the data distribution is very different. 

Replacing our greedy peer selection strategy with random peer sampling decreases performance, as our method intelligently selects the best peers while random peer sampling may select peers that have very different data distributions. 
What is surprising is that AlphaEdge with greedy peer sampling even works better than NoPS without peer sampling (each edge communicates with every other edge) on the AirQuality dataset. This is because the dataset has 35 edges. Properly learning the aggregation weights for all $35$ neighbors for each edge can be difficult. However, with our greedy peer sampling method, each edge figures out the most informative $K=5$ neighbors and focuses on learning the aggregation weights for these $5$ neighbors which is much easier that than learning  $35$ weights. Also, by intelligently selecting the most informative neighbors, we are also removing possible adversarial peers whose information might hurt local model performance. Therefore, our greedy peer sampling method can not only reduce communication cost but also potentially improve performance.

Figure~\ref{fig:scores_vs_t} shows how the score of one typical edge changes over time for each dataset. The x-axis is normalized time obtained by dividing the current time by the total time span, i.e. $\frac{\text{current time}}{\text{total time span}}$. Our method is mostly on the top. Generally, when time increases, the score increases, this is because the model has seen more data batches so it has better performance. The scores can also fluctuate a lot especially on Dataset-B and AirQuality. This is likely due to significant data distribution shift on these two datasets.  

\subsection{Variation of Learning Environment}
In this section, we test the robustness of our method to different variations of the learning environment. In addition, we consider the best performing alternative method for each component in Table~\ref{tbl:performance}, FedWeight and RandPS. Note we do not consider NoPS as it does not perform peer selection and involves quadratic amount of communication which is not scalable.
Again note that on Dataset-A and Dataset-B, since we are allowing 5 neighbors, RandPS will have the same result as AlphaEdge.

\subsubsection{Delayed feedback}
We first consider a variation of our online learning setting where the ground-truth feedback is not immediately available but has a time delay. This is the case in less real-time online item recommendation scenarios. For example, when an user is recommended with a list of news items, the user might go through the items one by one, so that the user feedback (whether the user clicked on each item or not) is only available after a time delay. 
Let $T_{\text{batch}}$ denote the time duration of one data batch (which can be different on different dataset) and $\alpha_{\text{delay}} T_{\text{batch}}$ denote the time delay of the feedback. We vary $\alpha_{\text{delay}}$ in values $[0, 5, 10, 20]$ and show results in Figure~\ref{fig:delay}. 

First, AlphaEdge outperforms the other two methods on all datasets under all $\alpha_{\text{delay}}$. Again, note on Dataset-A and Dataset-B, RandPS is identical to AlphaEdge. Second, as the time delay $\alpha_{\text{delay}}$ increases, the performance of all methods decreases on Dataset-A, Dataset-B, and AirQuality. 
This is because these dataset can have real-time data shift. 
For example, for the online-store datasets (Dataset-A and Dataset-B),
some items might suddenly become popular due to trends on social media. 
If there is a delay in incorporating the latest feedback into the model, there is a delay picking up the new data trend, causing lower performance.
Similarly, air quality trend can drift in real time due to sudden weather change like sudden rain and storm.
Without immediate ground-truth  feedback, the model is not able to adapt to data shift as fast. On Movielens, the performance of all methods almost do not change when $\alpha_{\text{delay}}$ increases. This is because there is little real-time data shift on Movielens as users' ratings toward a movie do not change dramatically.

\subsubsection{Asynchronous edges}
\begin{figure}[!htb]
\vspace{-3mm}
    \centering
    \includegraphics[width=0.5\linewidth]{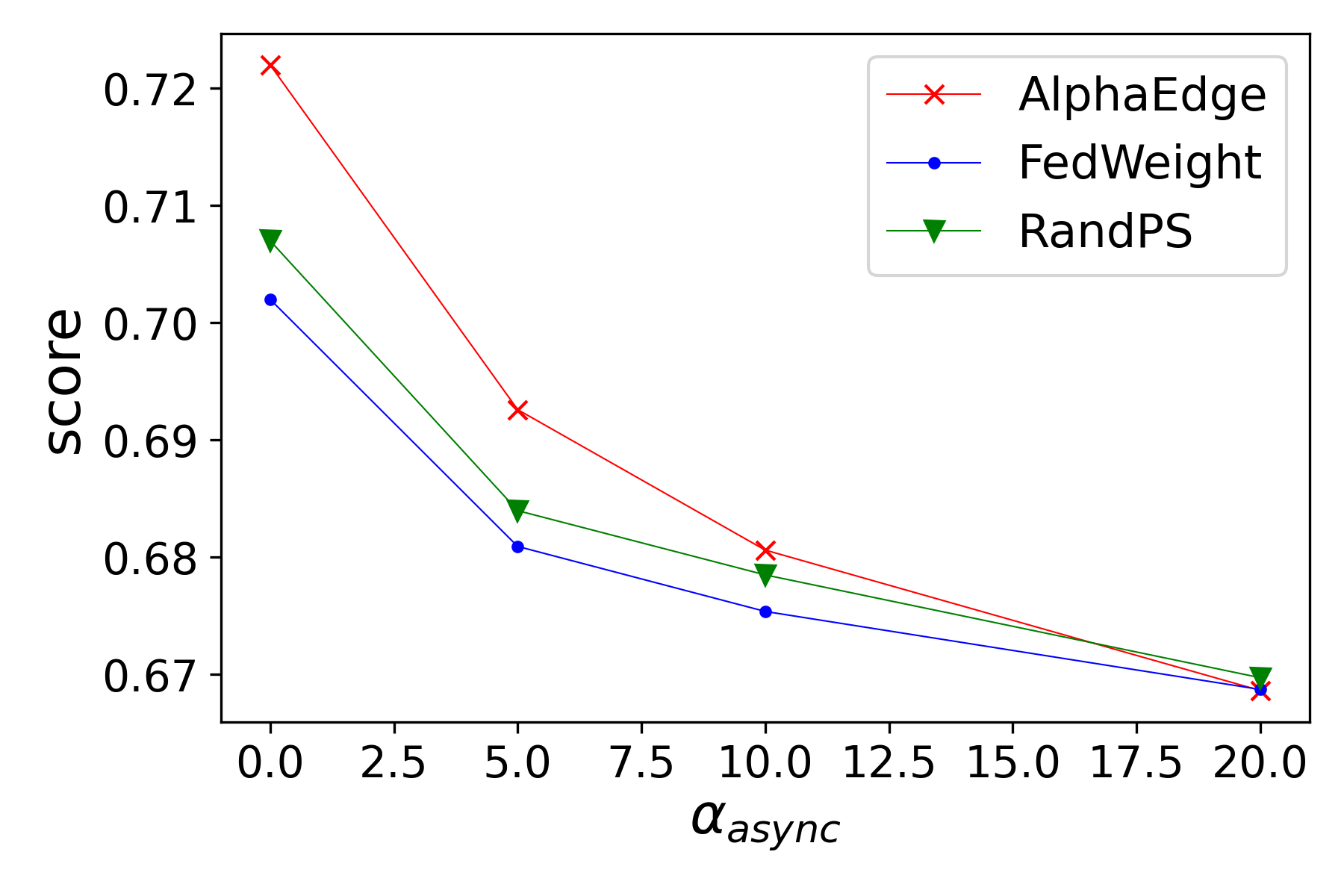}
    \vspace{-2mm}
    \caption{Averaged score over all datasets vs $\alpha_{\text{async}}$.}
    \vspace{-3mm}
    \label{fig:async}
\end{figure}

The edges may not be synchronized. Specifically, for edge $e^i$, during aggregation at timestamp $t$, the fetched model $M^j$ from neighbor $e^j$ maybe from a previous timestamp $t'$ with $t_{\text{async}} = t-t'>0$ due to different reasons like slow network connection. 
Let $\alpha_{\text{async}}= t_{\text{async}}/T_{\text{agg}}$ denote the amount of asynchrony. 
To simulate this scenario, for every neighbor of every edge, we vary $\alpha_{\text{async}}$ in values $[0, 5, 10, 20]$. Since all datasets have a similar trend, we only show the averaged scores over all datasets in Figure~\ref{fig:async}. As $\alpha_{\text{async}}$ increases, the score of all methods decreases as each edge gets "out-dated" information from its neighbors which is less helpful. The advantage of AlphaEdge also decreases as there is little gain by doing a more careful aggregation on "out-dated" information from neighbors.

\subsubsection{Network Connection Down}
In some cases, connection to some edges might be completely down. We simulate this scenario by randomly dropping connections of the edges. Specifically, for each edge, we randomly select $\alpha_{\text{drop}}$ percent of time to make it unavailable to other edges. 
When an edge wants to aggregate its local model with models of its $K$ neighbors, one neighbor happens to be not available, this neighbor will not participate during aggregation. This means on average each time $K-\alpha_{\text{drop}}K$ neighbors participate during aggregation.
For our method AlphaEdge, if one neighbor is down but become available again after a few rounds, its aggregation weight is still initialized as the weight learned in the last time when it was up. 
In practice it is unlikely that a server edge is often down, so we set $\alpha_{\text{drop}}=0.25$. 
We show the results in Table~\ref{tbl:network_down}. The results are similar to those in Table~\ref{tbl:performance}, except that the scores are a little lower. This is because for each edge some neighbors are not available for some time and the edge would not be able to incorporate information from these neighbors. 

\begin{table}[!htb]
\vspace{-2mm}
\centering
\caption{Performance with $25\%$ of time network down.}
\begin{tabular}{|l|l|l|l|}
\hline
           & FedWeight & RandPS & AlphaEdge      \\ \hline
Dataset-A  & 0.669     & \textbf{0.692}  & \textbf{0.692} \\ \hline
Dataset-B  & 0.792     & \textbf{0.812}  & \textbf{0.812} \\ \hline
Movielens  & 0.682     & 0.675 & \textbf{0.702} \\ \hline
AirQuality & 0.621     & 0.634  & \textbf{0.656} \\ \hline
Avg.       & 0.691     & 0.703  & \textbf{0.716} \\ \hline
\end{tabular}
\label{tbl:network_down}
\vspace{-2mm}
\end{table}

\subsubsection{Adversarial edges}
Our method naturally handles the scenario of adversarial edges. We simulate this scenario by flipping the labels of randomly chosen $25\%$ edges which serves as the adversarial edges. Specifically, for the chosen adversarial edges of the three recommendation datasets, we change every label from 1 to be 0 (or 0 to be 1); for  the chosen adversarial edges of the regression dataset, since the label $y$ is continuous, we "flip" each label $y$ as $y_{\text{max}}+y_{\text{min}}-y$ where $y_{\text{max}}$ and $y_{\text{min}}$ are the maximum label value and minimum label value among all data points. 
We report the scores averaged on the normal edges (those without label flipping) in Table~\ref{tbl:adv}.

\begin{table}[!htb]
\vspace{-2mm}
\centering
\caption{Performance with $25\%$ adversarial edges.}
\begin{tabular}{|l|l|l|l|}
\hline
           & FedWeight & RandPS & AlphaEdge      \\ \hline
Dataset-A  & 0.416     & \textbf{0.647}  & \textbf{0.647} \\ \hline
Dataset-B  & 0.588     & \textbf{0.758}  & \textbf{0.758} \\ \hline
Movielens  & 0.657     & 0.668  & \textbf{0.705} \\ \hline
AirQuality & 0.596     & 0.608  & \textbf{0.663} \\ \hline
Avg.       & 0.564     & 0.670  & \textbf{0.693} \\ \hline
\end{tabular}
\label{tbl:adv}
\vspace{-2mm}
\end{table}

First, the performance of AlphaEdge is lower than its performance in Table~\ref{tbl:performance}. This is because each edge would have less helpful neighbors if its original most helpful neighbors happen to be chosen as the adversarial edges. Second, the performance of FedWeight and RandPS decreases significantly. This is because FedWeight is not able to adapt the weights intelligently to put less/zero weights on the adversarial edges. For RandPS, although it has the AlphaEdge's component of learning aggregation weights, it is not exploiting the weights to select the most helpful peers, so normal edges still select the adversarial edges as neighbors, which sabotages local model performance (even though RandPS might have learned to put smaller weights on these adversarial neighbors).

\subsection{Different hyper-parameters}
The most important hyper-parameters are the number of epochs between two aggregation steps $E$ (this controls aggregation frequency) and batch size $B$ of the streaming data batches. We vary the two hyper parameters respectively. 

\subsubsection{Number of epochs till next aggregation.}
\begin{figure}
\centering
\begin{subfigure}{.5\linewidth}
  \centering
  \includegraphics[width=1.0\linewidth]{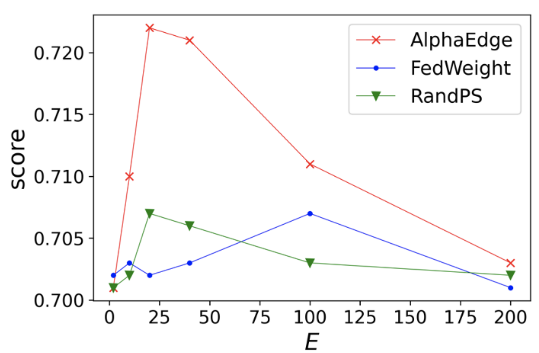}
  \caption{}
  \label{fig:tagg_1}
\end{subfigure}%
\begin{subfigure}{.5\linewidth}
  \centering
  \includegraphics[width=1.0\linewidth]{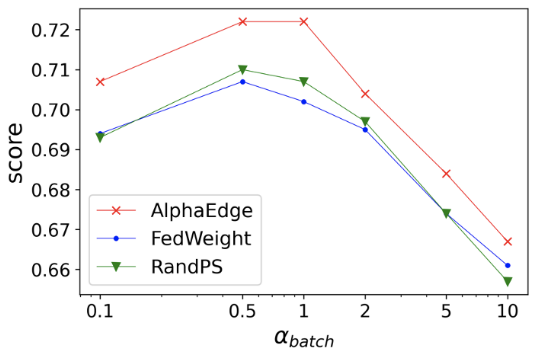}
  \caption{}
  \label{fig:tagg_2}
\end{subfigure}
\caption{Averaged score over all datasets vs (a) $E$ and (b) $\alpha_{\text{batch}}$}
\label{fig:test}
\vspace{-6mm}
\end{figure}

We vary $E$ in values [2, 5, 10, 20, 40, 100, 200]. The results are shown in Figure~\ref{fig:tagg_1}. As $E$ increases, the performance of all methods first increases and then decreases (though they peak at different values). The decreasing part can be easily explained as when $E$ increases to be extremely large all methods become identical to NoAgg which has a score of $0.699$. When $E$ is relatively small, performance increases as $E$ increases. We note a similar trend is observed in existing federated learning methods~\cite{mcmahan2017communication}. The conjecture is that with a larger $E$ the local models are more different so aggregating them has a regularization benefit~\cite{mcmahan2017communication}. For our method, another reason is that we use the data batch at the aggregation step to learn the aggregation weights but not the model parameters, i.e. overall $\frac{1}{E}$ of the data is not used for learning model parameters. When $E$ is very small, a considerable amount of data is not used for learning the model parameters, so increasing $E$ is beneficial.

\subsubsection{Data batch size.}
Let $B_0$ denote our default batch size (which is 50 for the two smaller datasets and 500 for the two large datasets). We vary batch size $B$ by changing $\alpha_{\text{batch}}=\frac{B}{B_0}$ in values [0.1, 0.5, 1, 2, 10]. We show the results in Figure~\ref{fig:tagg_2}. For all methods, as the batch size increases, the averaged score first increases and then decreases. This is because when the batch size is very small, the gradient is very noisy so that the model may deviate from the global minimum; When the batch size is very large, we have less number of gradient updates, so that the model is not optimized well.  
Overall, AlphaEdge always performs better than the other methods.

\section{Conclusion}
In this work, we propose a new learning setting Decentralized Personalized Online Federated Learning targeting at applications that require reliability, personalization, online learning and privacy at the same time. We identify two technical challenges in this learning setting: how to perform aggregation and how to select neighbors for each edge/client. We propose to directly learn an aggregation by optimizing the local model performance with respect to the aggregation weights, and then use the learned aggregation weights to select neighbors for each edge/client. We verify the effectiveness and robustness of our proposed method on four real-world item recommendation and regression datasets.

\bibliographystyle{IEEEtran}
\bibliography{main.bib}

\end{document}